\documentclass[runningheads,a4paper]{llncs}

\usepackage{amssymb}
\usepackage{graphicx}
\usepackage{booktabs}
\usepackage{tabularx}
\usepackage{multirow}
\usepackage{rotating}
\usepackage{verbatim}
\usepackage{cite}
\usepackage{subfigure}

\usepackage{array}
\newcolumntype{L}{>{\arraybackslash}m{\columnwidth}}

\usepackage{color}

\usepackage{url}
\urldef{\mailsa}\path|{alfred.hofmann, ursula.barth, ingrid.haas, frank.holzwarth,|
\urldef{\mailsb}\path|anna.kramer, leonie.kunz, christine.reiss, nicole.sator,|
\urldef{\mailsc}\path|erika.siebert-cole, peter.strasser, lncs}@springer.com|
\urldef{\mymail}\path|{xirong,qjin}@ruc.edu.cn|

\newcommand{\etal}{\textit{et al.}~}
\newcommand{\keywords}[1]{\par\addvspace\baselineskip
\noindent\keywordname\enspace\ignorespaces#1}

\begin{document}

\mainmatter  

\title{Improving Image Captioning by Concept-based Sentence Reranking}

\titlerunning{Improving Image Captioning by Concept-based Sentence Reranking}

%
%
\author{Xirong Li and Qin Jin\thanks{Corresponding author (qjin@ruc.edu.cn)}}
\authorrunning{Li \emph{et al.}}

\institute{Key Lab of DEKE, Renmin University of China\\ 
Multimedia Computing Lab, Renmin University of China\\
}

%
%

\toctitle{Lecture Notes in Computer Science}
\tocauthor{Authors' Instructions}
\maketitle

\begin{abstract}

This paper describes our winning entry in the ImageCLEF 2015 image sentence generation task.
We improve Google's CNN-LSTM model by introducing \emph{concept-based sentence reranking},
a data-driven approach which exploits the large amounts of concept-level annotations on Flickr.
Different from previous usage of concept detection that is tailored to specific image captioning models,
the propose approach reranks predicted sentences in terms of their matches with detected concepts,
essentially treating the underlying model as a black box.
This property makes the approach applicable to a number of existing solutions.
We also experiment with fine tuning on the deep language model, 
which improves the performance further.
Scoring METEOR of 0.1875 on the ImageCLEF 2015 test set, 
our system outperforms the runner-up (METEOR of 0.1687) with a clear margin.
\end{abstract}

\section{Introduction} \label{sec:intro}

\begin{figure}[tb!]
\centering
\subfigure[~]{
                \includegraphics[width=0.9\textwidth]{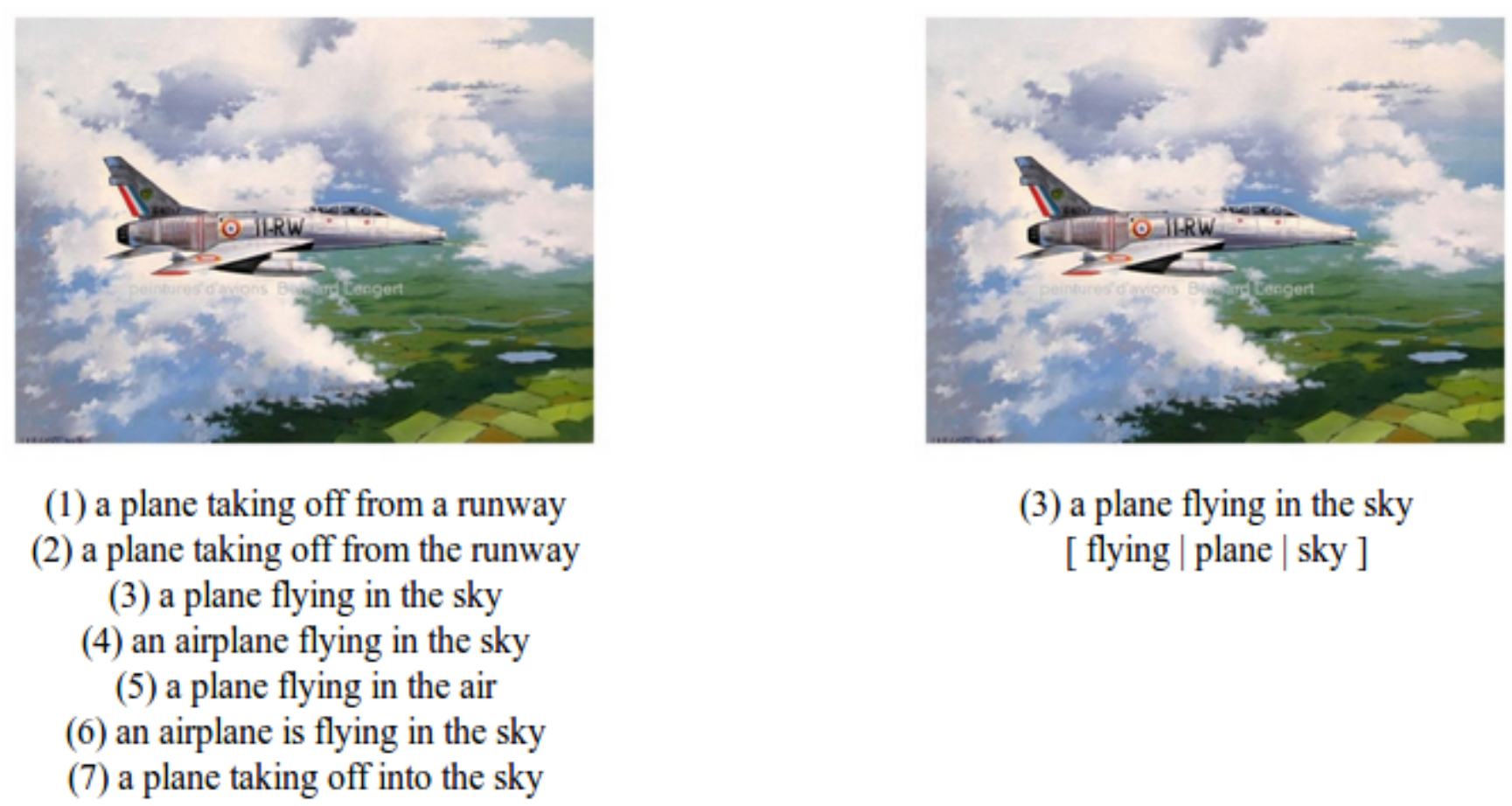}
                \label{fig:rerank-plane}
}
\subfigure[~]{
                \includegraphics[width=0.9\textwidth]{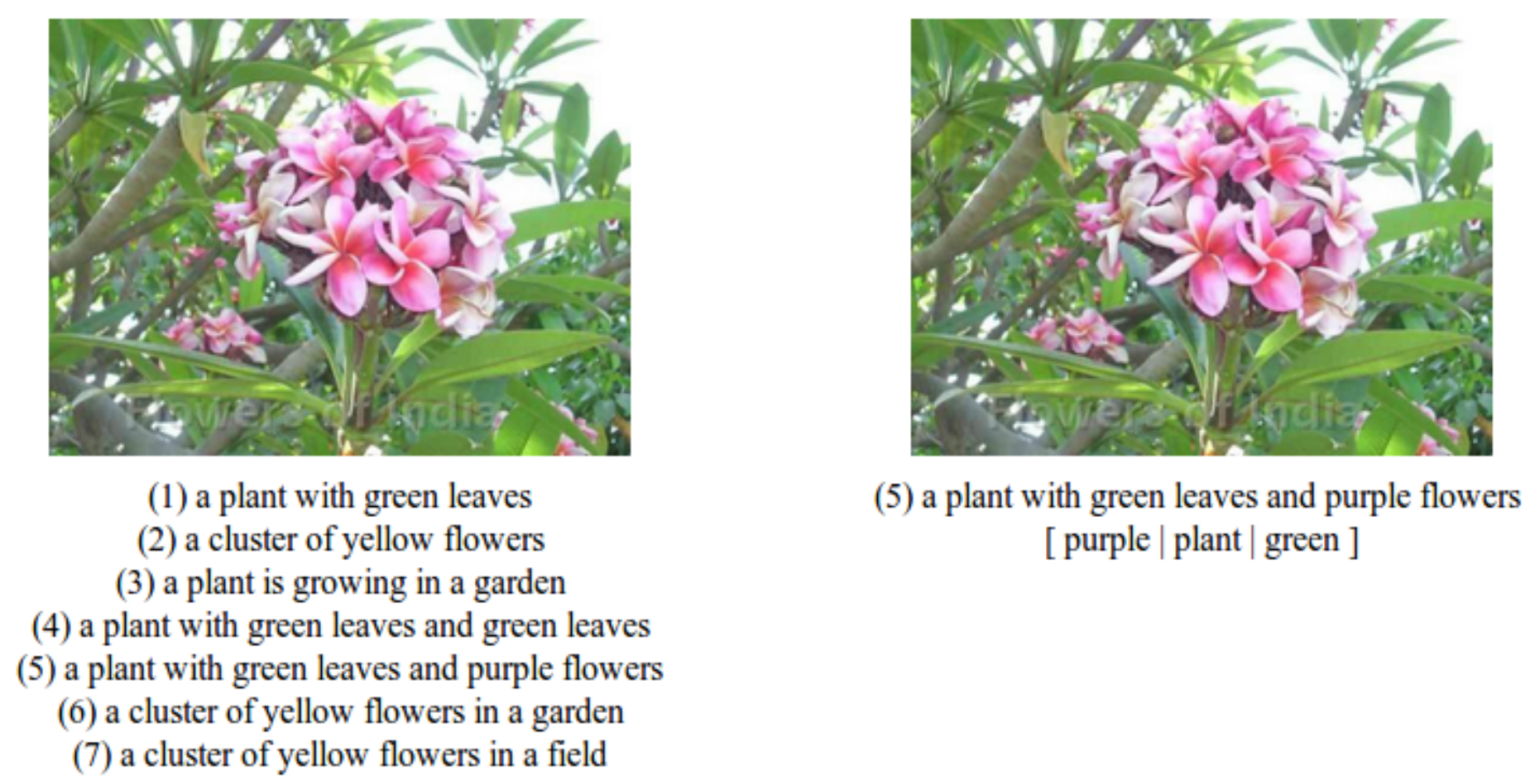}
                \label{fig:rerank-flower}
}                
\caption{
\textbf{Examples illustrating concept-based sentence re-ranking for improving image captioning}.
Candidate sentences generated by a CNN-LSTM image captioning model are shown in descending order on the left side, 
while the chosen sentences are shown on the right side.
In square brackets are concepts predicted by the neighbor voting algorithm \cite{tmm2009-tagrel}.
The candidate sentence best matching the predicted concepts are chosen as the final description.
} \label{fig:rerank}
\end{figure}

In this paper we tackle the challenging task of image captioning.
Given an unlabeled image, the task is to automatically generate a natural language sentence that describes main entities and events present in the image. See Fig. \ref{fig:rerank} for some example sentences.


There has been a considerable progress on the topic in the last few years,
thanks to powerful image representation derived from deep convolutional neural networks (CNN) \cite{nips2012-hinton} 
and trainable recurrent neural networks (RNN) capable of modeling long term dependency in natural language \cite{nips2014-seq2seq}.
Joint models of CNN and RNN have demonstrated quite promising results for image captioning \cite{google-show-tell,iclr15-mao-mrnn,cvpr2015-neuraltalk}.

Since the number of model parameters to be optimized is at a million scale,
what also matters are the growing amounts of training images associated with manually written descriptions, 
e.g., Flickr8k \cite{flickr8k}, Flickr30k \cite{flickr30k}, MSCOCO \cite{mscoco} with more than 100k images, and Flickr8k-CN \cite{flickr8kcn} as a bilingual extension of Flickr8k. 
Even though the number of captioned images has increased from a few thousands to over 100k, 
sentence-level annotations remain in shortage when compared to concept-level annotations.
For instance, one can easily obtain one million learning examples of `dog' from Flickr\footnote{Over 6 million images tagged with `dog' on Flickr, \url{https://www.flickr.com/search/?tags=dog}, retrieved on April-29-2016.}. 
How to exploit such large amounts of (noisy) concept-level annotations to improve image captioning is important.

Works on utilizing concept detection for image captioning exist, but are tailored to specific models. 
For instance, in the work by Fang \etal \cite{cvpr2015-mscaption}, concept detection results are used as part of input features for a maximum entropy language model.
Gong \etal \cite{eccv2014-sentence-embedding} leverage Flickr data to improve image and text embedding within a kernel Canonical Correlation Analysis framework.
These works are inapplicable to other models, e.g., the popular CNN + LSTM architecture.

\medskip

\textbf{Contributions of this work}. 
For improving image captioning by exploiting concept-level annotations, 
we propose concept-based sentence reranking.
The proposed technique treats the underlying sentence generation model as a black box, making it applicable to a number of existing solutions.
Moreover, by deriving concept detectors from Flickr images, better image descriptions are predicted with no extra cost of manual annotation.
In addition,  we show that a fine tuning on a trained deep language model further improves the performance.
Putting all this together, our system is the winning entry in the ImageCLEF 2015 image sentence generation task \cite{ruc-clef2015,Gilbert15_CLEF}.

\section{Related Work} \label{sec:related}

Depending on whether candidate sentences are given in advance, 
we see two lines of research, namely retrieval approaches and generative approaches.

The retrieval approaches sort a predefined set of candidate sentences in terms of their relevance with respect to a given image, 
and then selects the top ranked sentence to annotate the image \cite{flickr8k,eccv2014-sentence-embedding,nips13devise}.
To compute cross-media relevance between images and sentences, representing them in a common space is a prerequisite. 
In \cite{flickr8k}, this space is derived by Kernel Canonical Correlation Analysis,
while normalized linear Canonical Correlation Analysis is used in \cite{eccv2014-sentence-embedding} for scaling to larger training sets.
In the DeViSE model \cite{nips13devise}, the common space is formed by a pretrained \emph{Word2Vec} \cite{word2vec},
where the embedding vector of a sentence is obtained by averaging over the vectors of its words.
Compared to bag-of-words used in \cite{flickr8k,eccv2014-sentence-embedding},
the use of \emph{Word2Vec} enables DeViSE to handle a much lager vocabulary.
In \cite{word2visualvec}, the common space is implemented as a visual CNN feature space in order to better capture visual and semantic similarity.
One strength of of the retrieval approaches is that a predicted sentence is syntactically and grammatically correct. 
However, they lack the ability to generate novel sentences.

The generative approaches typically consist of two main components, i.e., image encoding and sentence decoding \cite{google-show-tell,cvpr2015-neuraltalk,iclr15-mao-mrnn}.
For image encoding, a pretrained deep convolutional neural network (CNN) is employed to project a specific image into a visual feature space.
For cross-media comparison, images and words are embedded into a latent common space before they are fed to a 
recurrent neural network (RNN), which eventually outputs a sequence of words as the caption.
Variants of RNN have been investigated, 
including multi-modal RNN \cite{iclr15-mao-mrnn}, bidirectional RNN \cite{cvpr2015-neuraltalk}, and Long Short-Term Memory (LSTM) \cite{google-show-tell}.
Recent benchmark evaluations,  e.g., the MSCOCO  Captioning Challenge \cite{mscoco} and the ImageCLEF 2015 image sentence generation task \cite{Gilbert15_CLEF}, have demonstrated outstanding performance of LSTM based solutions \cite{google-show-tell,ruc-clef2015}.

\section{Our Approach} \label{sec:approach}

We aim to improve the performance of an existing image captioning model  by re-ranking its generated sentences in terms of concept detection results.
More concretely, given an image, the model first generates $k$ best sentences.
Each sentence, indicated by $hypoSent$, is associated with a confidence score, denoted as $sentenceScore(hypoSent)$.
Meanwhile, suppose we have access to a concept detection system that predicts $m$ concepts deemed to be relevant with respect to the given image.
If a detected concept appears in a $hypoSent$, we call it a matched concept.
Our assumption is that a sentence covering more matched concepts is more likely to be a good description of the image.
Hence, the final caption shall be the sentence that maximizes the prediction confidence score and the concept matches. 
The new sentence confidence score is therefore computed as a linear combination of these two factors, i.e.,
\begin{equation} \label{eq:sent-reranking}
newScore(hypoSent) = \theta \cdot concScore(hypoSent) + (1 - \theta) \cdot sentScore (hypoSent),
\end{equation}
where $concScore(hypoSent)$ is the averaged confidence score of all the matched concepts in the hypothesis sentence,
and $\theta \in [0,1]$ is a trade-off parameter.
Fig. \ref{fig:rerank} showcases some examples of how this simple strategy helps produce better image descriptions.

The proposed approach treats the image captioning component as a black box, meaning any captioning model can be applied in theory.
In this work we adopt Google's CNN-LSTM framework \cite{google-show-tell} for its state-of-the-art performance.
The framework is described in Section \ref{ssec:google-model}.
We then discuss in Section \ref{ssec:fine-tuning} how to adapt the model to new target data,
followed by our choices of concept detection in Section \ref{ssec:concept-detection}.

\subsection{The CNN-LSTM model for sentence generation} \label{ssec:google-model}

To generate the hypothesis sentences for a specific image, we employ the CNN-LSTM model proposed by Vinyals \etal \cite{google-show-tell}.
At the heart of the model is an LSTM based recurrent neural network, 
which computes the posterior probability of a (novel) sentence conditioned on the image.

The network is trained by maximum likelihood estimation using many pairs of image and sentence.
As an image and a sentence are of different modalities and thus not directly comparable,
an image projection matrix and a word projection matrix are deployed to embed the image and the sentence from their initial representations into a common space before feeding them into the network.
We use a CNN feature, i.e., the last fully connected layer of a pretrained 16-layer VGGNet \cite{vgg},
as the initial representation of an image.
Each word in a sentence is represented by a one-hot vector.

In the sentence generation stage, a hypothesis sentence of the input image is generated as a sequence of words in a greedy manner.
In particular, the CNN vector of the input image, after projection, is fed into the LSTM network to initialize its memory units.
The posterior probability over all words is re-estimated based on the memory status and the words already been chosen in previous iterations.
Consequently, the word with the highest probability is selected, while its embedding vector will be fed into the LSTM network in the next iteration. The generation process stops once a predefined STOP token is chosen.
Similar to \cite{google-show-tell}, we use beam search to obtain a list of hypothesis sentences.

\subsection{Model adaptation by fine tuning} \label{ssec:fine-tuning}

We use stochastic gradient descent to train the LSTM based sentence generation model.
Such an incremental learning strategy makes it relatively ease to adjust an existing model with respect to novel data.
In particular, we leverage a transfer learning strategy similar to the ones used in the context of object recognition for fine tuning a CNN model \cite{eccv2014-finetuning}. That is, re-train the LSTM model using a novel training dataset but with a relatively lower learning rate.

\subsection{Concept Detection} \label{ssec:concept-detection}

Notice that the output of the CNN model corresponds to the 1,000 labels defined in the Large Scale Visual Recognition Challenge \cite{ILSVRCarxiv14}. We do not directly use this as concept detection results, because the labels are often over specific to be overlapped with words used in image captions.
Instead, we consider two methods for concept detection, i.e., neighbor voting \cite{tmm2009-tagrel} and hierarchical semantic embedding \cite{sigir2015-hierse}, both of which are capable of learning from large amounts of weakly labeled Flickr images, 
and thus predict words used in daily life.

\textbf{Neighbor voting} (NeiVote) \cite{tmm2009-tagrel}. Given an image, this method first retrieves a set of neighbor images visually close to the given image,
and then count the occurrence of a specific concept in textual annotations associated with the neighbors. 
The concepts are sorted in descending order by their occurrence and the top $m$ ranked ones are preserved.
Despite its simplicity, a recent comparison \cite{csur2016-tagsurvey} shows that the method remains competitive when compared to more complicated alternatives. 
The visual distance between images are computed in terms of the Euclidean distance between their CNN features.

\textbf{Hierarchical Semantic Embedding} (HierSE) \cite{sigir2015-hierse}. 
Developed in the context of zero-shot image tagging, HierSE casts concept detection into cross-media relevance computation.
In particular, HierSE embeds both concept and image into a \emph{Word2Vec} space.
The embedding vector of the concept is obtained by convex combination of the embedding vectors of the concept and its ancestors tracing back to
the root in WordNet. 
The embedding vector of an image is obtained by convex combination of the embedding vectors of the top ten labels predicted by the CNN model.
Consequently, the image-concept relevance score is computed as the cosine similarity between the corresponding embedding vectors.
The top $m$ concepts with the largest scores are preserved.

\begin{figure}[tb!]
\centering
\noindent\includegraphics[width=\columnwidth]{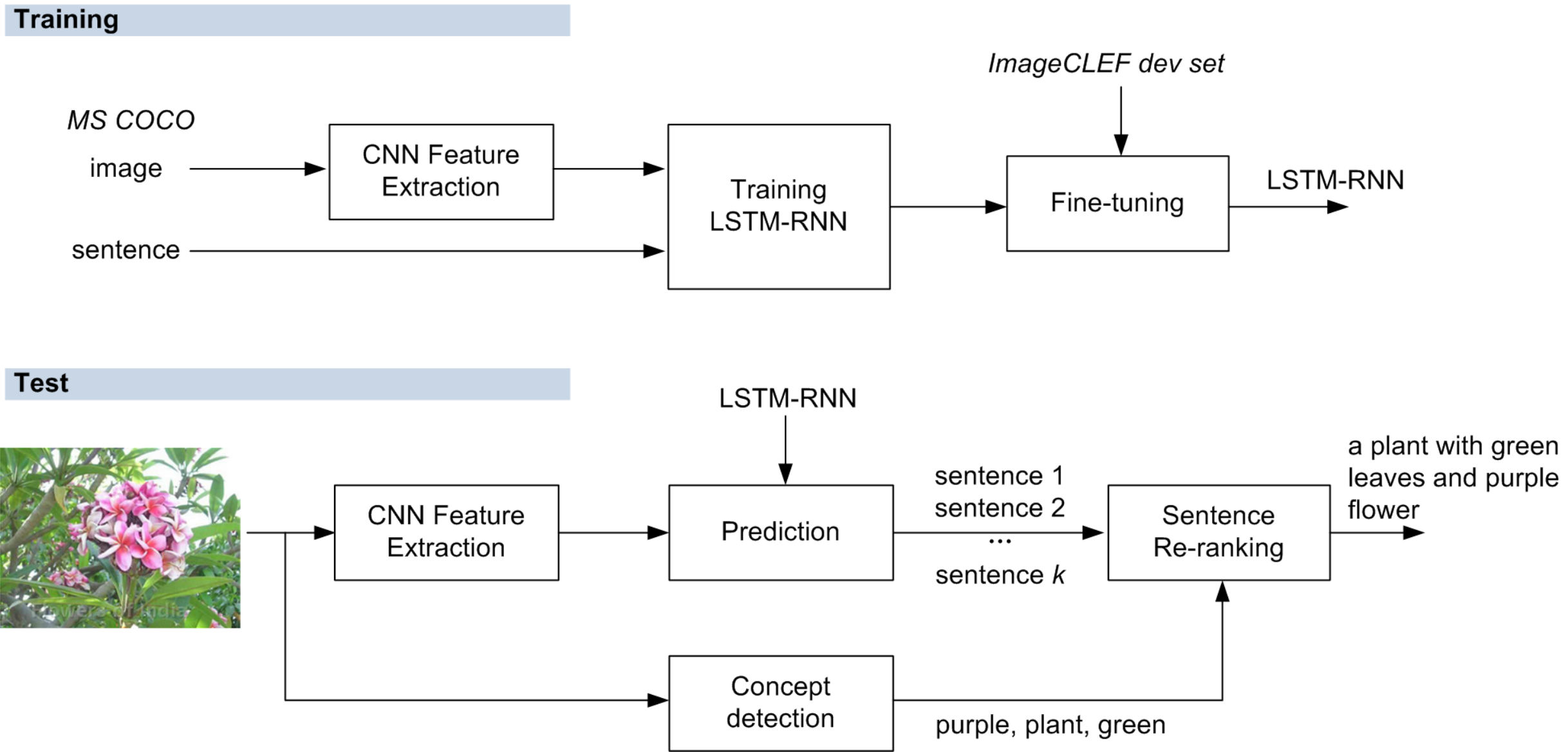}
\caption{\textbf{An illustration of the proposed solution}.
We improve the popular CNN + LSTM architecture through 1) model adaption by fine tuning and 2) sentence reranking using auto-detected visual concepts.}\label{fig:framework}
\end{figure}

A conceptual diagram of the proposed approach is given in Fig. \ref{fig:framework}.

\section{Evaluation} \label{sec:eval}


To verify the effectiveness of our approach, we participated in the ImageCLEF 2015 image sentence generation task \cite{Gilbert15_CLEF}.
In this task, participants were asked to generate sentence-level textual descriptions for 500k images crawled from the web,
from which the task organizers selected a subset of 3,070 images for performance assessment.
Notice that we have no access to ground-truth annotations, as the test set is used for blind testing by the organizers only.

In ImageCLEF 2015, a dev set of 2k images with manually written captions are provided for system development.
The number of captions per image ranges from 5 to 51 per image, with a mean of 9.5 and a median of 8 descriptions.
In order to tune the weighting parameter $\theta$ in Eq. \ref{eq:sent-reranking}, 
we randomly split the dev set into three disjoint subsets, i.e., 1,600 images for training, 200 images for validation and the remaining 200 images for internal test.

\subsection{Experiment 1. The influence of fine tuning}

We compare two runs. One run is to train the sentence generation model on the ImageCLEF dev set.
The other run is to first train the model on the  MSCOCO dataset \cite{mscoco}, and then fine tune it on the ImageCLEF dev set using a lower learning rate.
Following the protocol \cite{Gilbert15_CLEF}, we use the METEOR score to assess the performance of image captioning.
As shown in the first two rows of Table \ref{tab:compare-runs},
the fine-tuned model, with METEOR of 0.1759, performs better.

\subsection{Experiment 2. The effect of concept-based sentence reranking}

Given detection results either from NeiVote or from HierSE, 
we apply concept-based sentence reranking on sentences generated by the CNN-LSTM models with and without fine tuning, respectively.
For NeiVote, we retrieve neighbor images from a collection of 2 million Flickr images.
For HierSE, we adopt an existing \emph{Word2Vec} model pre-trained on Flickr tags \cite{sigir2015-hierse}.

\begin{table}[tb!]
  \renewcommand{\arraystretch}{1.2}
  \centering
  \caption{\textbf{Performance of our solution under varied settings on the ImageCLEF 2015 image sentence generation task}.}
  \label{tab:compare-runs}%
  \scalebox{1}{
     \begin{tabular}{@{}|l|r|r|r|@{}}

    \toprule

    \textbf{Training data} & \textbf{Data for fine tuning} & \textbf{Concept detection} & \textbf{METEOR score} \\
    \midrule
    ImageCLEF dev   & --  & --  & 0.1659 \\
    \midrule
    MSCOCO &  ImageCLEF dev & --  & 0.1759 \\
    \midrule
    ImageCLEF dev   & --   & HierSE & 0.1781 \\
    \midrule
    ImageCLEF dev   & --   & NeiVote & 0.1806 \\
    \midrule
    MSCOCO & ImageCLEF dev & HierSE & 0.1684  \\
    \midrule
    MSCOCO & ImageCLEF dev & NeiVote & \textbf{0.1875} \\
    \bottomrule
    \end{tabular}%
  }
\end{table}%

As we see from Table \ref{tab:compare-runs}, the performance of the model (without fine tuning) consistently improves after sentence reranking,
with the METEOR score increases from 0.1659 to 0.1781 (HierSE) and 0.1806 (NeiVote), respectively.
For the fine-tuned model, using HierSE for concept detection causes some performance drop,
while NeiVote remains effective, reaching the best METEOR score of 0.1875.
The results justify the effectiveness of concept-based sentence reranking, 
and NeiVote appears to be a better choice for concept detection in this context.

\begin{figure}[!tb]
\centering
\includegraphics[width=0.9\textwidth]{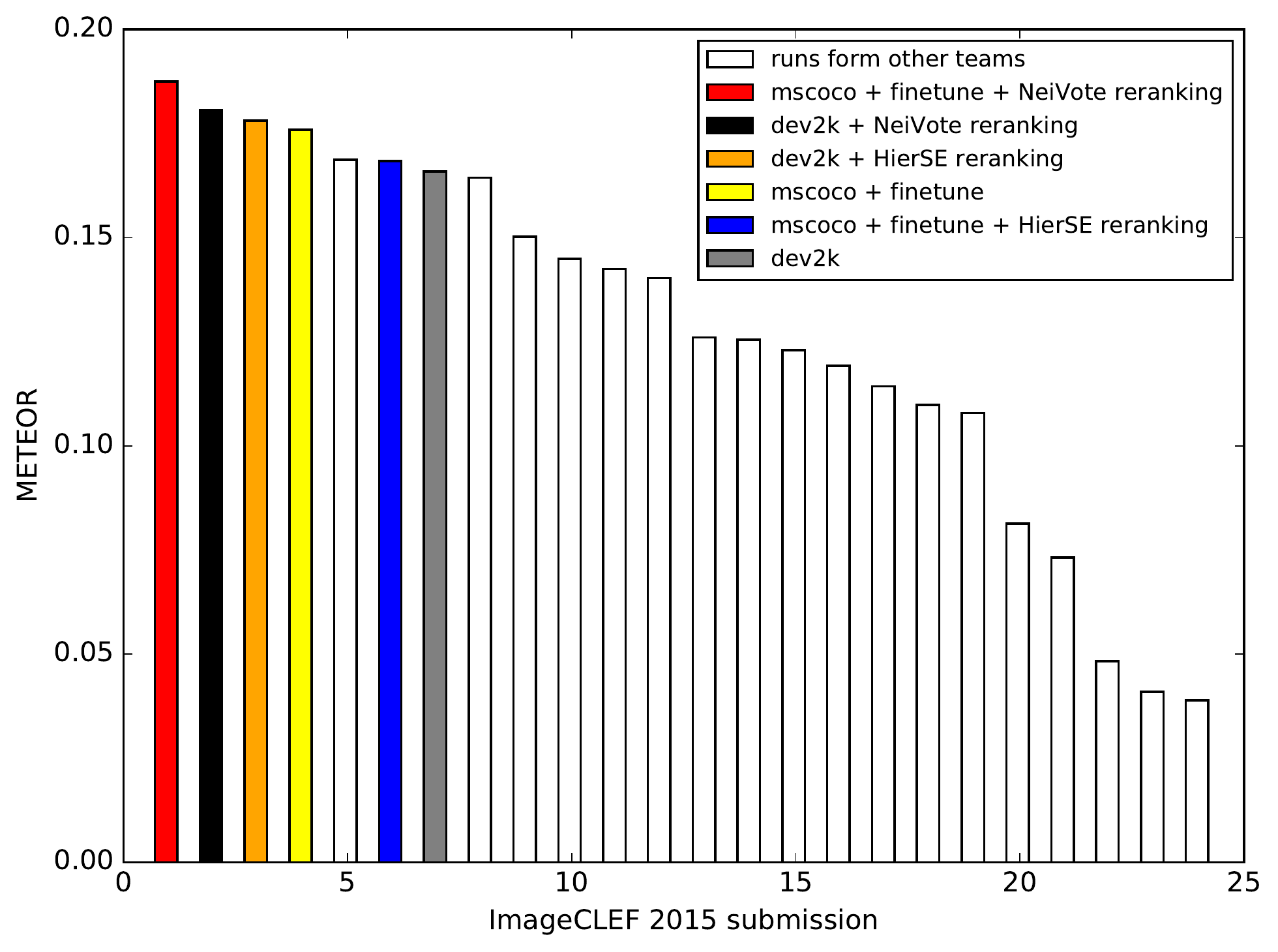}
\caption{\textbf{Comparing with submissions from other teams in the ImageCLEF 2015 image sentence generation task}. 
Our results given varied settings are highlighted in color bars.
The submissions have been sorted in descending order according to their METEOR scores.} \label{fig:versus-others}
\end{figure}

\begin{figure}[!tb]
\centering
\noindent\includegraphics[width=\columnwidth]{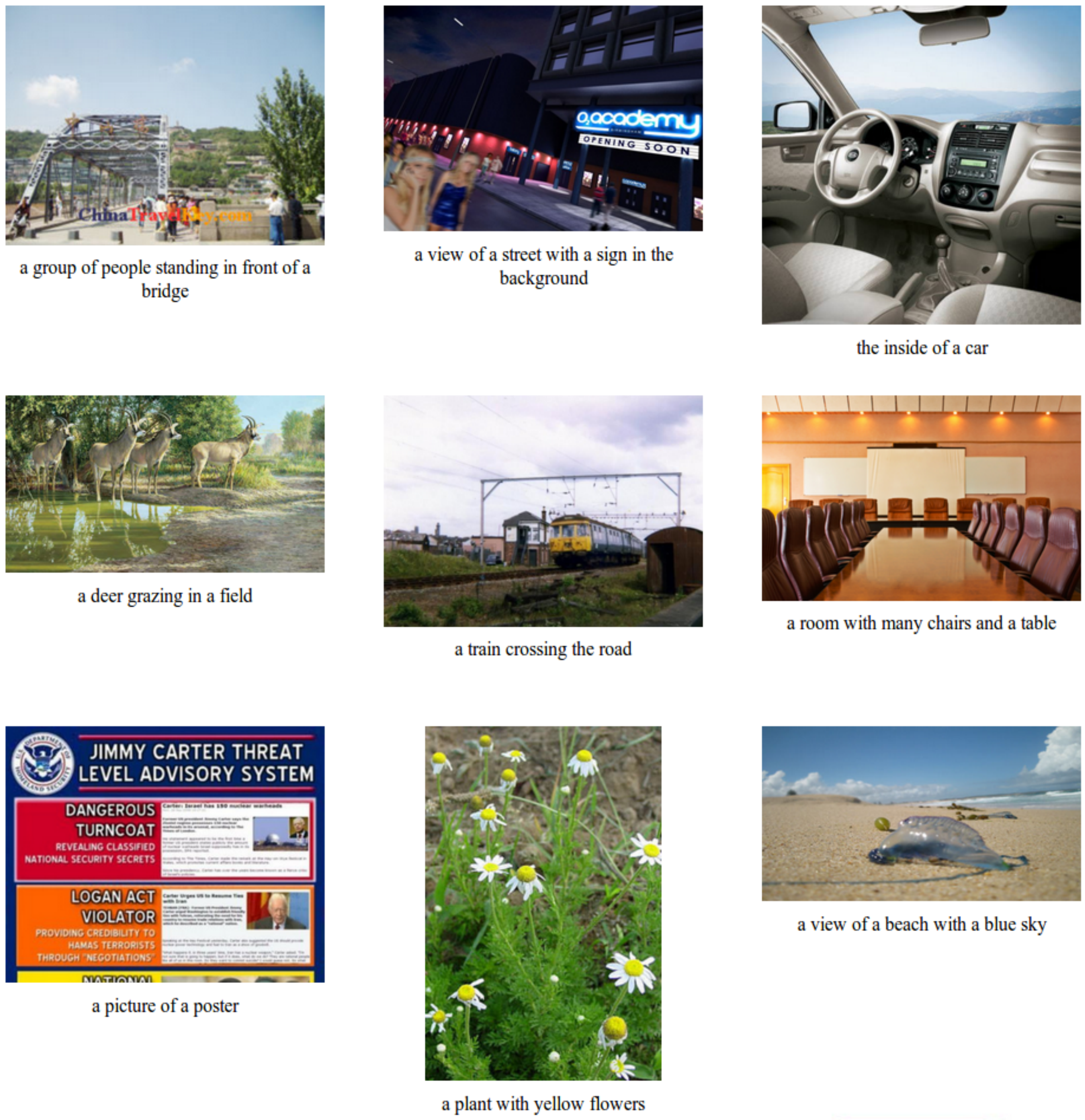}
\caption{\textbf{Some test images with sentences generated by the proposed system}. 
The images are hand picked that showing our system performs well.}\label{fig:task2-output}
\end{figure}

Fig. \ref{fig:versus-others} plots the performance of our system in the context of all submissions in ImageCLEf 2015,
showing our leading position in the evaluation.
Image captioning results of some selected examples are given in Fig. \ref{fig:task2-output}.
The quantitative and qualitative results demonstrate the potential of the proposed concept-based sentence reranking.

\section{Conclusions} \label{sec:conc}

We present in this paper \emph{concept-based sentence reranking}, a data-driven approach to improve image captioning.
As demonstrated by our participation in the ImageCLEF 2015 benchmark evaluation,
the proposed approach is found to be effective for improving the popular CNN-LSTM image captioning model.
In essence the improvement is gained by exploiting the large amount of noisy concept-level annotations associated with Flickr images.
In addition, fine tuning on the deep language model helps its generalization to novel data.

\subsubsection*{Acknowledgements.}
The authors are grateful to the ImageCLEF coordinators for the benchmark organization efforts \cite{Villegas15_CLEF,Gilbert15_CLEF}.

\bibliographystyle{splncs}
\bibliography{sent-rerank}

\end{document}